\documentclass[letterpaper, 10 pt, conference]{ieeeconf}  % Comment this line out if you need a4paper

\IEEEoverridecommandlockouts                              % This command is only needed if 
\usepackage{amsfonts}
\usepackage{bm}
\usepackage{makecell}
\usepackage{dsfont}
\usepackage{booktabs} % for professional tables
\usepackage[linesnumbered,ruled,norelsize]{algorithm2e}
\usepackage{hyperref}
\usepackage{amsmath}
\newcount\Comments % 0 suppresses notes to selves in text
\usepackage{array,multirow,graphicx}
\usepackage{color}
\usepackage{wrapfig}
\definecolor{darkgreen}{rgb}{0,0.5,0}
\definecolor{purple}{rgb}{1,0,1}
\newcommand{\kibitz}[2]{\ifnum\Comments=1\textcolor{#1}{#2}\fi}
\usepackage{stfloats}
\usepackage{tabularx}
\let\labelindent\relax  % Fixes issue with ieeestyle redefining \labelindent: https://tex.stackexchange.com/questions/170772/command-labelindent-already-defined
\usepackage{enumitem}
\usepackage{subcaption}
\usepackage{graphicx}
\usepackage{siunitx}
\usepackage{float}
\title{\LARGE \bf
Mobile Manipulation Leveraging Multiple Views
}

\author{David Watkins-Valls$^{1}$, Peter K Allen$^{1}$, Henrique Maia$^{1}$, Madhavan Seshadri$^{1}$, Jonathan Sanabria$^{1}$, \\ Nicholas Waytowich$^{2}$, and Jacob Varley$^{3}$% <-this % stops a space
\thanks{$^{1}$Department of Computer Science, Columbia University, New York, NY, USA, {\tt\small\{davidwatkins,allen,henrique\}@cs.columbia.edu}, {\tt\small \{ms5945,js5425\}@columbia.edu}}%
\thanks{$^2$U.S. Army Research Laboratory, Baltimore, MD, USA. {\tt\small nicholas.r.waytowich.civ@mail.mil}}%
\thanks{$^3$Robotics at Google. {\tt\small jakevarley@google.com}}
\thanks{This research was sponsored by the Army Research Laboratory and was accomplished under Cooperative Agreement Number W911NF-18-2-0244. The views and conclusions contained in this document are those of the authors and should not be interpreted as representing the official policies, either expressed or implied, of the Army Research Laboratory or the U.S. Government. The U.S. Government is authorized to reproduce and distribute reprints for Government purposes notwithstanding any copyright notation herein.}
}

\begin{document}

\maketitle
\thispagestyle{empty}
\pagestyle{empty}

\begin{abstract}
While both navigation and manipulation are challenging topics in isolation, many tasks require the ability to both navigate and manipulate in concert. To this end, we propose a mobile manipulation system that leverages novel navigation and shape completion methods to manipulate an object with a mobile robot. Our system utilizes uncertainty in the initial estimation of a manipulation target to calculate a predicted next-best-view. Without the need of localization, the robot then uses the predicted panoramic view at the next-best-view location to navigate to the desired location, capture a second view of the object, create a new model that predicts the shape of object more accurately than a single image alone, and uses this model for grasp planning. We show that the system is highly effective for mobile manipulation tasks through simulation experiments using real world data, as well as ablations on each component of our system.
\end{abstract}
\section{Introduction}
\label{sec:Introduction}

Fully autonomous mobile manipulation has long been an important goal in robotics, with a particular focus on such wide-ranging applications as manufacturing, warehousing, construction, and household assistance~\cite{kalashnikov2018qt, mahler2019learning, jacobus2015automated}. Mobile manipulation encompasses a sequence of robot navigation, object detection, view planning, grasp planning, and grasp execution that makes it a challenging task. The challenge of mobility becomes exacerbated when the task environment is dangerous to be explored by a human. Recent works addressing mobile manipulation aim to map, traverse, and grasp in unknown and partially observable environments~\cite{schwarz2017nimbro,orsag2017dexterous,wang2020multi}. Active perception to set up a goal based on some current belief to achieve an action is a good model for how to potentially solve this problem~\cite{bajcsy2018revisiting}. 

Navigating accurately and efficiently in an environment is a crucial first step to achieving autonomous mobile manipulation. Traditional position and mapping focused algorithms include a Simultaneous Localization and Mapping (SLAM)~\cite{dissanayake2001solution} technique to plan a collision free path. Such techniques, while effective in mapping and localizing in an unknown environment, are sensitive to odometry errors and noise. Increasingly, reinforcement learning based techniques have also been used to solve navigation in complex environments~\cite{zhu2017target,francis2020long,mirowski2016learning}; however, reinforcement learning uses sparse rewards and requires an extremely large amount of training episodes to achieve a good navigation model.

% \begin{figure}[t]
%     \centering {
%         \includegraphics[trim={5cm 0 5cm 0},clip,width=0.95\linewidth]{imgs/cover.pdf}
%     }
%     \caption{Two unregistered views are superior to a single view, especially when a single view does not give proper depth information. We show a series of different improvements in completion quality for objects not seen at runtime. Better completion accuracy means higher likelihood of grasping. } \label{fig:coverfigure}
% \end{figure}

\begin{figure}[t]
    \centering {
        \includegraphics[width=0.95\linewidth]{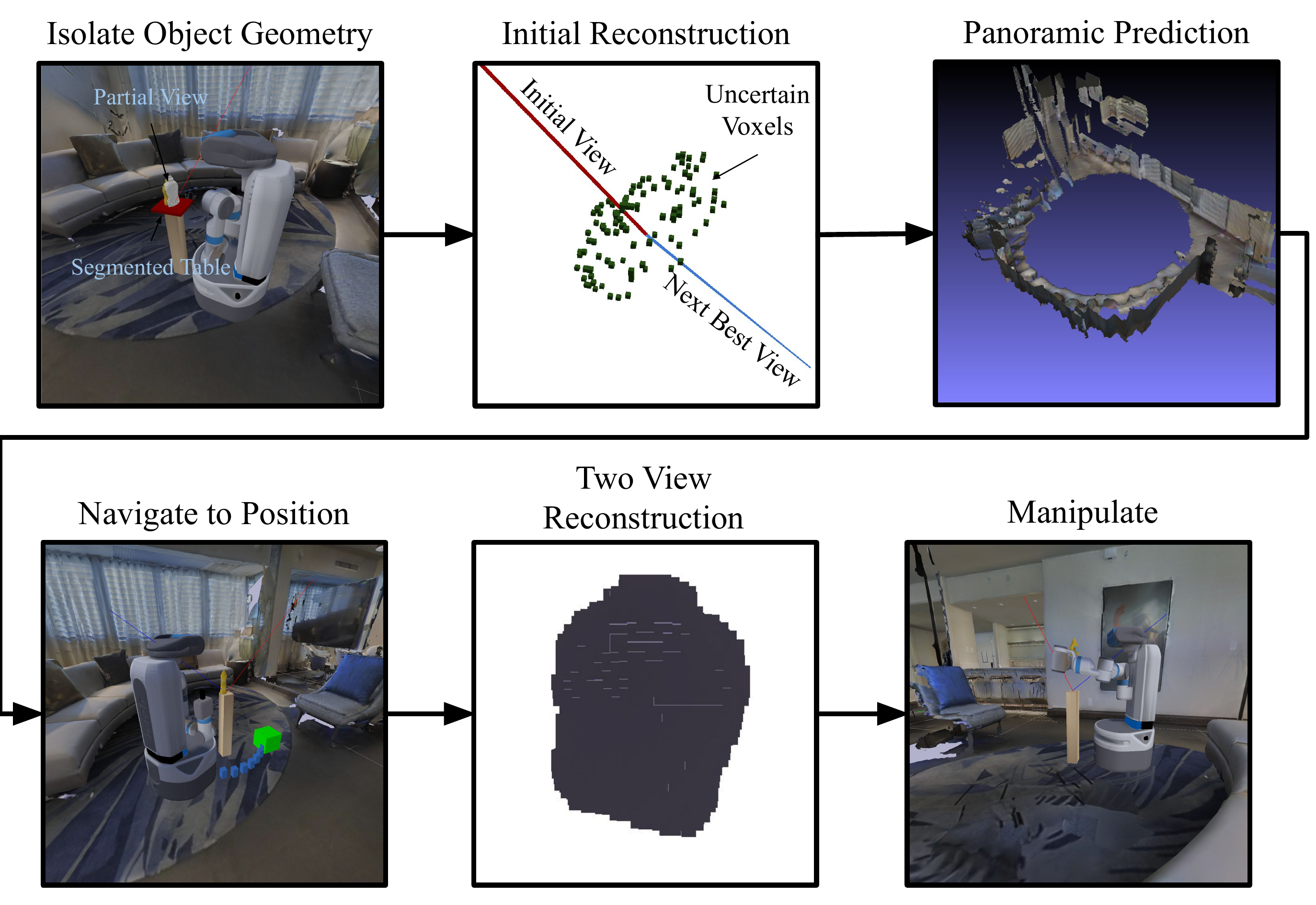}
    }
    \caption{The mobile manipulation system combines navigation to a target goal with shape understanding through a series of discrete stages to decide how to acquire additional information about the object for manipulation. It is a shape completion system that can utilize two unregistered views to get a better model of the target object. The system is described in more detail in Section \ref{sec:Methods}.} \label{fig:overview}
\end{figure}

%In previous work, we have enabled robots to navigate through learned environments to a panoramic target goal~\cite{watkins2019learning} and in another work we have enabled manipulation through shape completion~\cite{watkins2019multi}.
Overall, we present a system that fuses two views to help estimation: based on the first view we determine a goal pose from which we can get the best second view, we generate a panoramic navigation goal for this second view, and lastly reuse the navigation ability to get the next-best-view. With a completed object shape we are then able to optimize our approach from the combined views and achieve more accurate grasps. Notably, we find that by utilizing a learning-based navigation pipeline the robotic agent can execute an efficient plan to navigate to the next-best-view without relying on absolute position at runtime. This system can generalize to real-world environments as it is trained on real world data via the YCB~\cite{calli2015ycb} and Matterport 3D~\cite{Matterport3D} datasets.

The main contribution of our work is an end-to-end mobile manipulation system leveraging multiple views for grasping shown in Figure \ref{fig:overview}. The system segments out the point cloud of the object at the goal location and uses a shape completion pipeline to predict the object's geometry. Using this partial view it finds a novel next-best-view to compute a second image to take of the object and uses our learned navigation system to navigate to the second view. Our use of a mobile robot enables us to find a next-best-view of the object that will refine our initial hypothesis of the object the most.  Afterwards, it captures a second view and uses a CNN trained on two unregistered views, a two-view CNN, to complete the object shape. Using this refined object hypothesis the system plans a grasp on the object to lift the object.  We test our system on a simulated robot agent in environments captured from the real world. 
Towards this goal, our contributions can be summarized as follows:

\begin{itemize}
    % \item An end-to-end system for mobile manipulation of household graspable objects utilizing novel learning algorithms
    \item An algorithm that takes an initial shape completion estimate of the manipulation target that uses voxel grid occupancy thresholding to plan the next-best-view
    \item An algorithm to generate panoramic targets from nearby scans and leverage our existing learned navigation method~\cite{watkins2019learning} to arrive at nearby locations
    \item A learned two-view shape completion method that given the initial and next-best-view creates a more accurate reconstruction for robotic manipulation without the need for registering the views in the same coordinate system
    \item Ablation studies demonstrating the utility of various components as well as overall system performance
    \item An open source dataset of trajectories, object point clouds, and object placements in real world scanned environments to reproduce our results (all accessible at \url{https://crlab.cs.columbia.edu/mobilemanipulation}).
    
\end{itemize}

\section{Related Work}
\label{sec:RelatedWork}

Mobile manipulation has been explored in a variety of contexts, but we restrict our comparison to household manipulation research. One of the first examples of mobile manipulation is HERB~\cite{HERB} allowing a robotic agent to select grasping targets and navigate through an environment. However, it required checkerboard localization and precise sensors to plan tasks. Nevertheless, this work helped form the basis for robotic mobile manipulators, such as separating  manipulation and navigation tasks. Some works have come out more recently advancing on this initial vision that utilize global localization and point cloud reconstruction in household environments~\cite{gofetch, DOMel2017, Wu2020}. 

Indoor map exploration provides a mobile robot with ample opportunity to navigate around the environment, but becomes increasingly difficult when deprived of sensory information. Several authors have proposed reinforcement learning techniques for this task~\cite{zhu2017target,anderson2018evaluation,bansal2020combining}. We share a common goal as Zhu et al.~\cite{zhu2017target} where the task is to navigate using images. Some assume the presence of an idealized global localization system where they train the bot to reach the goal location~\cite{anderson2018evaluation, lind2018deep, richter2017safe, relomogen}. Another aims to solve the problem of traversing unknown environments by combining model based control with learning based perception where the task is to come up with a set of waypoints leading up to the goal~\cite{bansal2020combining}. 

On approaching the goal, the mobile robot captures images nearby containing the object of interest. However, this image only represents the raw sensory data which is incomplete due to the field of view and the approach angle. Grasp planning with such minimal and incomplete information is a challenging task prone to failures. Accurate shapes of objects improves grasp planning and execution success rate. Several works have proposed a deep learning approach to produce predictive mesh representation of partial views for unseen objects through either 3D convolutions or graph convolutions for robotic grasping~\cite{varley2017shape, dai2017shape, litany2018deformable, watkins2019multi}. Other shape completion systems exist for household objects but not robotic grasping~\cite{Yang18, 3DNVS}. Several geometric solutions to object 3D modeling have been proposed as well~\cite{williams2006gaussian, krainin2010manipulator, krainin2011autonomous, hermann2016eye}. 

A mobile manipulation system also needs to leverage the ability to acquire additional information by moving a vision system. To find candidate next-best-views we need an algorithm to evaluate the quality of our current geometric understanding and the cost of acquiring additional views. A foundational work from Connolly did this by identifying positions of the sensor that will maximize data collected~\cite{connolly1985determination}. Many works came after Connolly et al. that utilized heuristic approaches to determine obstructions that would block future collected data~\cite{pito1995solution, Callieri2004RoboScanAA, chen2005vision, gomeznbvplanning}. A recent survey of next-best-view algorithms showed that Chen et al. had the best next-best-view performance~\cite{karaszewski2016assessment}. Another work from McGreavy et al.~\cite{mcgreavy2017next} looked at next-best-view planning using a cylindrical model to analyze the visibility of an initial object candidate and find an optimal view. Their model does not use a learned shape completion system and relies on knowing objects beforehand. Some newer works in shape segmentation are also applicable for next-best-view work as they allow for better segmentation for unseen object geometry~\cite{xie2021unseen}. Other work in occlusion based grasping has shown success in mapping voxels using registered views~\cite{kahn2015active}. One possible method to capture a next-best-view is to use an eye-in-hand camera as shown by Potthast et al.~\cite{Potthast2011NextBV}, however there are kinematic restrictions with respect to the workspace of the robot that prevent these views from being captured. Other works have presented object reconstruction under uncertainty that utilize an algorithmic approach to estimate the object's geometry while utilizing the odometry of the mobile robot~\cite{gomez2014}. 

\section{Methods}
\label{sec:Methods}

In a previous work, we created a learning-based navigation pipeline which does not rely on odometry, map, compass, or indoor position at runtime and is purely based on the visual input and an 8-image panoramic goal. Our method learns from expert trajectories generated using RGBD maps of real world environments. We showed our system was able to learn efficiently across multiple environments and form shorter paths than state of the art vision based navigation systems~\cite{watkins2019learning}. Our agent uses a policy model to move forward, left, or right through the environment using the current view and four previously captured RGBD images or history buffer. We train a separate goal checking model that reports whether the agent has reached the desired location. In this work, we leverage this navigation system along with a two-view shape completion system to accomplish mobile manipulation. 

\subsection{Isolate Object Geometry}
The robot captures a 8-image RGBD panoramic view of its current surroundings. This panorama is converted into a point cloud. Points within a boundary of $0.3m$ and $0.7m$ from the robot are filtered. We then search for planes in that cloud parallel to the base of the robot using RANSAC~\cite{fischler1981random}. Once the plane is found, we can then segment all points above this plane as the object of interest. Because our system is modular, any shape segmentation system is a valid replacement. Our results show that RANSAC is sufficient for isolated object geometry. With this partial view, the agent begins the \textit{Completion} stage. 

\subsection{Completion}
\label{sec:completion}
The point cloud generated by the \textit{Isolate Object Geometry} stage is used to perform an initial shape completion of the object. In a previous work, we developed a shape completion CNN architecture which took a partial voxelized view of the object and output a voxel grid hypothesis of occupancy scores where $1$ is filled and $0$ is empty with a decision boundary of $0.5$ or $v_{boundary}$. Our completion system is similar to our previous work in shape completion~\cite{watkins2019multi} except we do not directly turn this hypothesis into a mesh for grasping. Given that we have a mobile robot we can find a next-best-view of the object that will refine our initial hypothesis the most. 

\begin{figure*}[ht]
    \centering {
        \includegraphics[width=0.95\textwidth]{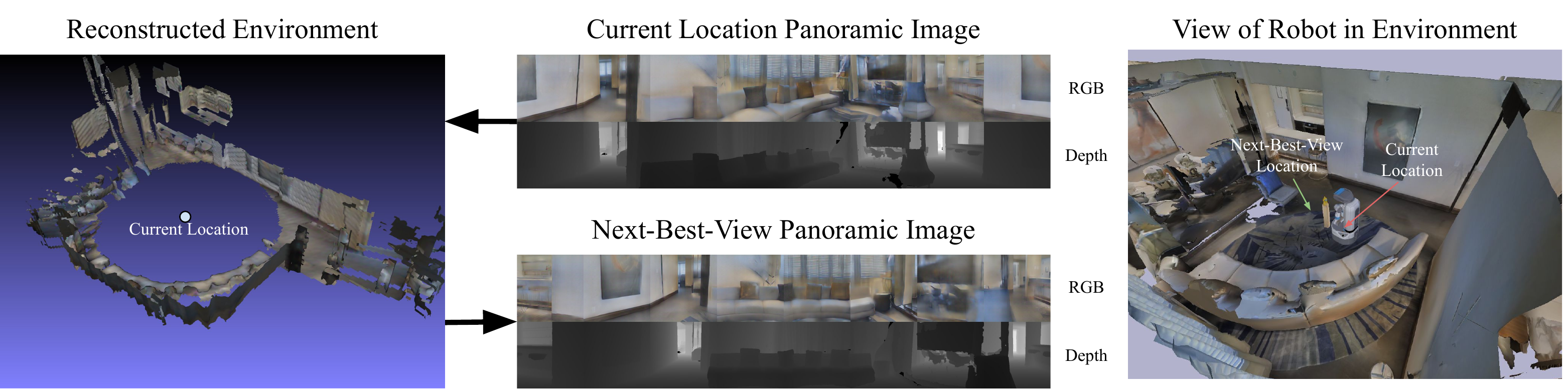}
    }
    \caption{We capture a panoramic image at the current location of the robot, which is then turned into a reconstruction of the nearby environment using Open3D~\cite{open3d}. We then render the target panoramic image at the Next-Best-View location in this reconstruction to navigate to it without localization. We show this predicted panorama at the Next-Best-View location here.  }\label{fig:predictedpanorama}
\end{figure*} 

\subsection{Next-Best-View}
\label{sec:next_best_view}
Voxels in this initial hypothesis that have an occupancy score of $0.5 \pm\epsilon$ are considered uncertain as they are close to the decision boundary and a second observation would be helpful to determine their occupancy. $\epsilon$ is the error bounds at $\pm0.025$. We calculated a bounds of $\epsilon$ to be $\pm0.025$ by evaluating the completion quality of various bounds and found that we had the best performance at $\pm0.025$. Ultimately, we want to capture as many of these points as possible in a secondary camera view while still being reachable by the robotic agent. Principle Component Analysis (PCA) solves this objective in a 3D space by taking the smallest component which is orthogonal to a 2D plane that best fits the data. This smallest component will capture the least variance and therefore observe the most voxels. This vector is calculated as shown in Equation (\ref{eq:pca}) where $\mathbf{X}$ is the set of uncertain voxels, $\mathbf{w}$ is the corresponding eigenvector, and $\hat{X}_k$ is the eigenvectors calculated by PCA.
% Appendix \ref{app:nbvprobability}
\begin{equation}
    \boldsymbol{v_{nbv}} = \arg\min \mathbf{\hat{X}}_k = \arg\min (\mathbf{X} - \sum_{s = 1}^{k - 1} \mathbf{X} \mathbf{w}_{(s)} \mathbf{w}_{(s)}^{\mathsf{T}}\label{eq:pca})
\end{equation}

\noindent
This next-best-view vector $v_{nbv}$ does not consider the height of the robot. We then extract the $(x, y)$ components of the vector, normalize the vector, and then multiply it by $d_{optimal}$ as an empirically optimal distance from the target object. We then assign a $z$ value of $0$ for the target position for the robotic agent to navigate towards. The target robot position $\boldsymbol{p_{target}}$ will then be $(x_{target}, y_{target}, 0)$ relative to the current position of the robot, $(0, 0, 0)$. We then calculate a viewing angle $\theta_{head}$ and height adjustment $h_{torso}$ for the robot agent to best align the camera with $\boldsymbol{v_{nbv}}$. The final computations of $\boldsymbol{p_{target}}, \theta_{head}, h_{torso}$ are then used in the \textit{Panoramic Prediction} stage.

\subsection{Panoramic Prediction}

To navigate to a goal location and capture the next-best-view, we utilize a panoramic target goal. We can utilize the panorama taken in the \textit{Isolate Object Geometry} stage to produce a mesh reconstruction on the initial point cloud of the surrounding environment. We used the Open3D~\cite{open3d} implementation of Bernardini's ball-pivoting reconstruction paper~\cite{bernardini1999ball}. Using this mesh, we can then predict the panorama from $\boldsymbol{p_{target}}$ by loading the mesh into a renderer and taking 8 RGBD views at equal $45^\circ$ intervals. This mesh will have holes in the RGB view. We then use Gibson's inpainting system called filler~\cite{igibson} to resolve any missing data in the predicted view. This filler system was used to generate each image in the training dataset for the \textit{Navigation} stage. Once we have this predicted panoramic view, we can then utilize our learned navigation system to locally navigate to this next view without localizing the agent. When the agent has arrived at this location the agent raises its torso to $h_{torso}$ and adjusts its camera to align with $\theta_{head}$. Once the agent has aligned itself with the next-best-view, the \textit{Two View Completion} stage starts.  An example predicted panorama is shown in Figure \ref{fig:predictedpanorama}.

\subsection{Two View Completion}

\begin{figure}[t]
\vspace{1mm}
    \centering {
        \includegraphics[width=0.95\linewidth]{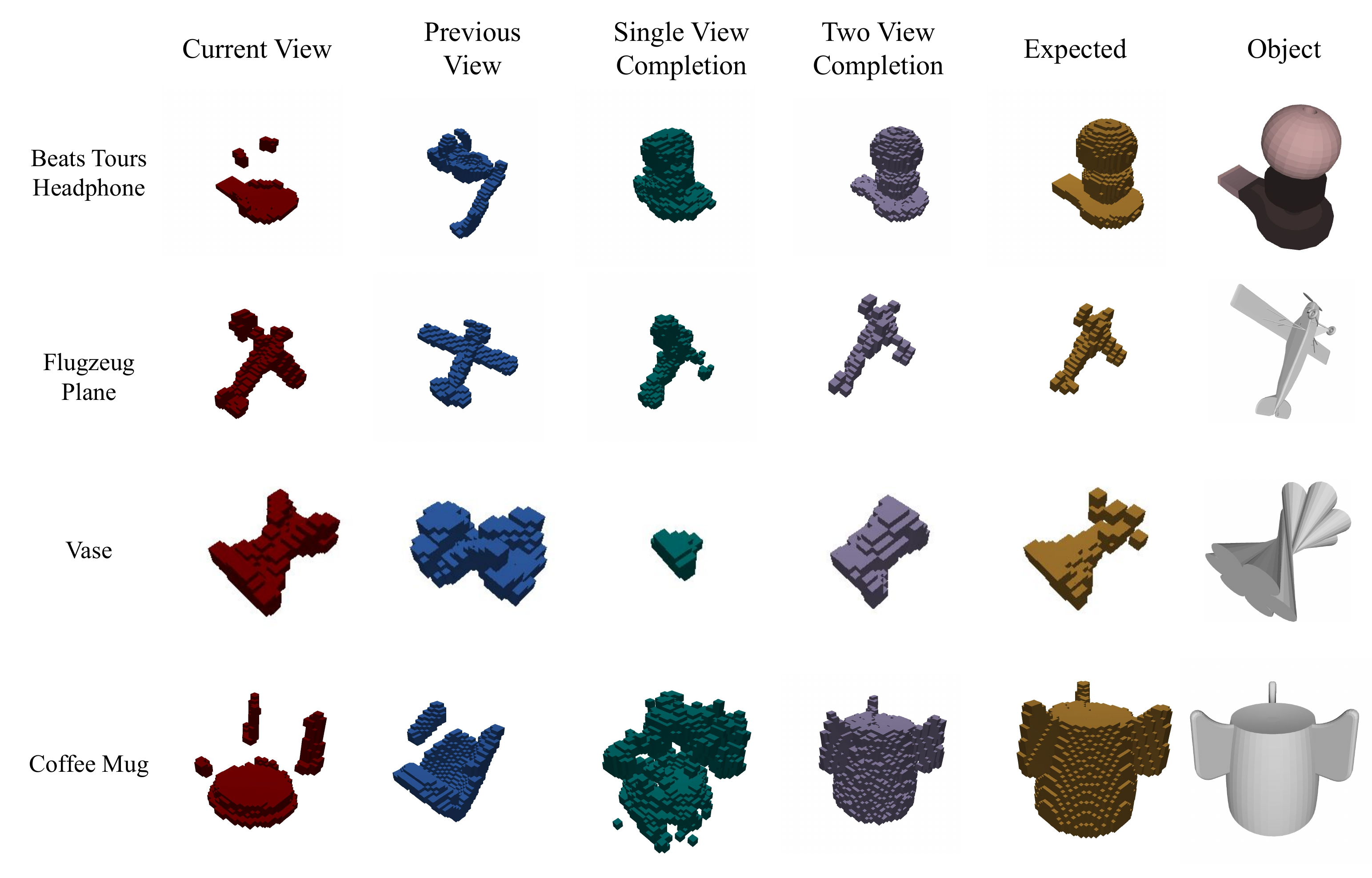}
    }
    \caption{Two view reconstruction inputs showing that the next-best-view is better than a single view. Current input (red), a single view reconstruction (green), the previous view of the object (blue), the two view completion (purple), and finally the ground truth mesh (yellow). These meshes were not observed during training for either network. All meshes are from the Challenge dataset. }\label{fig:completion_improvement}

\end{figure} 

Our previous shape completion work did not address the problem of two unregistered views of an object. In order to make sure the mobile robot does not need to localize at runtime, we designed a new CNN architecture that takes two views, the current and previous, that are both kept in the their original image frame. As input, it takes two voxelized partial views and outputs a voxelized reconstruction of the original object geometry. Similar to the \textit{Completion} stage, this model outputs a voxel grid of occupancy scores. We can now threshold the scores near $v_{boundary}$ and perform marching cubes~\cite{lorensen1987marching} to mesh the resultant voxel grid. In this format, we plan a manipulation task as described in the \textit{Manipulation} stage. Example completions of objects from our challenge dataset are shown in Figure \ref{fig:completion_improvement} which demonstrates that a next-best-view via PCA can offer large improvement over a single view. 

% \begin{figure}[t]
%     \centering {
%         \includegraphics[width=0.95\linewidth]{imgs/Two View CNN Architecture.pdf}
%     }
%     \caption{Shape completion can be performed utilizing two views via separate encoders and adding them together to form a dense embedding which is then decoded into the completion of the shape. The network can be modified to accept different types of inputs easily as long as they can be represented as 3D voxel grids, such as two unregistered views.   } \label{fig:twoviewarchitecture}
% \end{figure} 

The point cloud from the previous capture in its image frame and the current view of the object in its image frame are both voxelized into $40^3$ voxel grids. We then generate a hypothesis about the object's geometry in the current view image frame and produce a $40^3$ voxel grid of occupancy scores given the two views of the object where $1$ is filled and $0$ is unfilled. We turn this voxel grid certainties into a mesh by thresholding occupancy scores near $v_{boundary}$. We then mark any voxels that are known to be empty by tracing a ray from the camera vector to points in the voxel grid and any voxels along that line that are not occluded are marked as empty. Labeling voxels fixes any erroneous shapes on the visible side of the object. 

\subsubsection{Training}
To train this model we randomly sampled views from the same dataset used to train the single view system, keeping the meshes used for holdout the same. We had considered using a more intelligent view sampling method, such as the opposite view or using the next-best-view, for generating training data. We found that our method performed better when provided a series of random views instead of a heuristic. We trained the model on roughly $400k$ pairs over the $618$ meshes and $726$ views per mesh.

\subsubsection{CNN Architecture}
Our two view architecture is composed of two heads, each taking a $40^3$ voxelized view of the object. Each input is passed through $4^3$ 3D convolution layers $3$ times until it reaches a dense ReLU layer of size $5000$. Each dense layer is added together then passed through another dense ReLU layer of size $5000$. The final dense layer is passed through to a $64000$ size dense layer which is then reshaped into a $40^3$ reconstruction of the object geometry. This network is trained with the same cross-entropy loss. 
%The reconstruction network architecture is shown in Figure \ref{fig:twoviewarchitecture}.

%A description of the training schedule, figure of the CNN architecture, and implementation details are in Appendix A.2.  %\ref{app:two_view_completion}. 

\subsection{Manipulation}

Using this mesh, we plan a grasp on the mesh using GraspIt!~\cite{miller2004graspit} to get a series of grasp candidates. Each of these grasp candidates are given an associated volume quality and are filtered for volume quality above 0~\cite{ferrari1992planning}. Then we use MoveIt~\cite{sucan2013moveit} to plan the pick plan given each grasp and pick the trajectory with the smallest execution time. In order to generate this pick plan we model the table and a region above the object as obstacles to ensure the trajectory does not disturb the object. If no trajectories or grasps are valid the program terminates. 

\section{Experiments}
\label{sec:Experiments}
%For a description on how we chose these objects and table positions see Appendix C.1. % 

\subsection{Reconstruction Quality Tests}

\label{sec:reconstructionqualitymetrics}
Our first goal was to validate that getting two views would result in better reconstruction quality. There are three metrics that let us validate our hypothesis: \textbf{Jaccard similarity}, \textbf{Hausdorff distance}, and \textbf{grasp joint accuracy}. %, and \textbf{grasp success}. 
\begin{enumerate}
    \item \textbf{Jaccard similarity} We used Jaccard similarity to evaluate the similarity between a generated voxel occupancy grid and the ground truth. The Jaccard similarity between sets A and B is given by:
    \[
    J(A, B) = \dfrac{|A\cap B|}{|A\cup B|}
    \]
    The Jaccard similarity has a minimum value of 0 where A and B have no intersection and a maximum value of 1 where A and B are identical~\cite{jaccard}.
    \item \textbf{Hausdorff Quality} The Hausdorff distance is a one-direction metric computed by sampling points on one mesh and computing the distance of each sample point to its closest point on the other mesh. It is useful for determining how closely related two sets of points are~\cite{huttenlocher1993comparing}.
    \item \textbf{Grasp Joint Accuracy} We use GraspIt!~\cite{miller2004graspit} to generate a series of grasp candidates for each predicted mesh and then choose the one with the highest volume quality on that predicted mesh. We then use a simulated BarrettHand to execute the grasp within the GraspIt! simulator on the ground truth mesh and calculate the difference between the expected joint values and the realized joint values. 
    % \item \textbf{Grasp Success} In order to evaluate our framework's ability to enable grasp planning, the system was tested in simulation using the same set of completions. The use of simulation allowed for the quick planning and evaluation of at least $1$ grasp up to $10$. GraspIt!~\cite{miller2004graspit} was used to plan grasps on all of the completions of the objects by uniformly sampling different approach directions. These grasps were then executed not on the completed object but on the ground truth meshes in the Gibson simulator. The Fetch robot executed the grasp on the table in simulation and if the object was suspended for $5$ seconds of simulated time it was considered a successful trial. 
    
\end{enumerate}
% Appendix \ref{app:reconstructionqualitymetrics}

To evaluate the performance of our method, we have five test scenarios. A \textbf{Single View} reconstruction using only the current view of the object that is utilizing the same model architecture as used in Varley et al.~\cite{varley2017shape}. A \textbf{Same View} reconstruction using our two-view architecture but where the current view is passed in twice. This is to evaluate the performance benefit of using a larger model for shape completion. A \textbf{Two View (Random)} reconstruction using our two-view architecture but the two views are chosen randomly about the object. A  \textbf{Two View (Opposite)} reconstruction using our two-view architecture where the first view is random and the second view is chosen by capturing the view opposite to the first. A \textbf{Two View (Next-Best-View)} reconstruction using our two-view architecture where the first view is random and the second view is chosen by calculating the next-best-view as described in Section~\ref{sec:next_best_view}. A \textbf{Three View} reconstruction using a modified version of our two-view architecture where the previous two views are passed into a single encoder and the current view is passed into its own encoder. These three embeddings are then added together and decoded into a reconstruction of the object. 
    % \item \textbf{Ground Truth}: We show the performance of our grasp planner by showing the Ground truth model through our system. The other metrics will all be idealized. 

\begin{table*}[t]
\vspace{2mm}
\centering
% \parbox{.49\linewidth}{
\begin{subtable}{.48\textwidth}
    \centering
    \begin{tabular}{|c|c|c|c|} %c|}
    \hline
    %This is the header for the table
    \multicolumn{1}{|c|}{\begin{tabular}[c]{@{}c@{}}\textbf{Next-Best-View} \\  \textbf{Method}\end{tabular}} 
    & \multicolumn{1}{c|}{\begin{tabular}[c]{@{}c@{}}\textbf{Jaccard} \\  \textbf{}\end{tabular}} 
    & \multicolumn{1}{c|}{\begin{tabular}[c]{@{}c@{}}\textbf{Hausdorff} \\  \textbf{}\end{tabular}} 
    & \multicolumn{1}{c|}{\begin{tabular}[c]{@{}c@{}}\textbf{Grasp Joint} \\  \textbf{Error}\end{tabular}} \\
    % & \multicolumn{1}{c|}{\begin{tabular}[c]{@{}c@{}}\textbf{Grasp Success} \\  \textbf{Rate}\end{tabular}} \\ 
    \hline
    %This is the body for the table
        Single-View     & 0.782         & 6.573        & $4.52^\circ$ \\ \hline%         & 0.743      \\ \hline
    	Same-View       & 0.802         & 6.423        & $4.36^\circ$ \\ \hline%         & 0.752      \\ \hline
    	Random          & 0.818         & 6.251        & $4.14^\circ$ \\ \hline%         & 0.784      \\ \hline
     	Opposite        & 0.826         & 5.421        & $3.85^\circ$ \\ \hline%         & 0.853      \\ \hline
    	Next-Best-View  & \textbf{0.852} & \textbf{4.912} & $\mathbf{3.28^\circ}$\\ \hline% & \textbf{0.867} \\ \hline
    % 	Three-Views     & 0.868 & 4.782 & $3.12^\circ$ \\ \hline% & 0.881 \\ \hline
    %  	Ground Truth    & 1.0 & 0.0 & $0.0^\circ$      \\ \hline% & 0.924 \\ \hline
    \end{tabular}
    \caption{\textbf{YCB \& Grasp Dataset Test Object Results}}\label{tab:reconstruction_quality_tests}
\end{subtable}
% \hfill
% \parbox{.49\linewidth}{
\begin{subtable}{.48\textwidth}
    \centering
    \begin{tabular}{|c|c|c|c|}
    \hline
    %This is the header for the table
    \multicolumn{1}{|c|}{\begin{tabular}[c]{@{}c@{}}\textbf{Next-Best-View} \\  \textbf{Method}\end{tabular}} 
    & \multicolumn{1}{c|}{\begin{tabular}[c]{@{}c@{}}\textbf{Jaccard} \\  \textbf{}\end{tabular}} 
    & \multicolumn{1}{c|}{\begin{tabular}[c]{@{}c@{}}\textbf{Hausdorff} \\  \textbf{}\end{tabular}} 
    & \multicolumn{1}{c|}{\begin{tabular}[c]{@{}c@{}}\textbf{Grasp Joint} \\  \textbf{Error}\end{tabular}}
    \\
    % & \multicolumn{1}{c|}{\begin{tabular}[c]{@{}c@{}}\textbf{Grasp Success} \\  \textbf{Rate}\end{tabular}} \\ 
    \hline
    %This is the body for the table
        Single-View  & 0.648          & 7.340 & $5.31^\circ$\\ \hline %         & 0.584      \\ \hline
        % Single-View (second) & 0.644          & 7.332 & $5.29^\circ$\\ \hline %         & 0.579      \\ \hline
        Same-View            & 0.663          & 7.284 & $5.16^\circ$\\ \hline %         & 0.592      \\ \hline
        Random               & 0.753          & 6.418 & $4.85^\circ$\\ \hline %         & 0.758      \\ \hline
        Opposite        & 0.831         & 5.924        & $4.53^\circ$ \\ \hline%         & 0.853      \\ \hline
        Next-Best-View       & \textbf{0.866} & \textbf{5.341} & $\mathbf{4.24^\circ}$\\ \hline % & \textbf{0.878} \\ \hline
        % Ground Truth  & 1.0 & 0.0 %  & 0.924 \\ \hline
    \end{tabular}
    \caption{\textbf{Challenge Dataset Results}}\label{tab:reconstruction_quality_tests_challenge}
\end{subtable}
% }
\caption{Measuring the performance of the reconstruction quality of test and challenge meshes given a single-view, same-view, random new view, opposite-view, next-best-view, as described in Section \ref{sec:reconstructionqualitymetrics}. None of these meshes were seen during training. A higher Jaccard is better. A lower Hausdorff is better. A lower Grasp Joint Error on the 3 finger BarrettHand is better.}
\end{table*}

\begin{figure}[t]
    \centering {
        \includegraphics[width=0.95\linewidth]{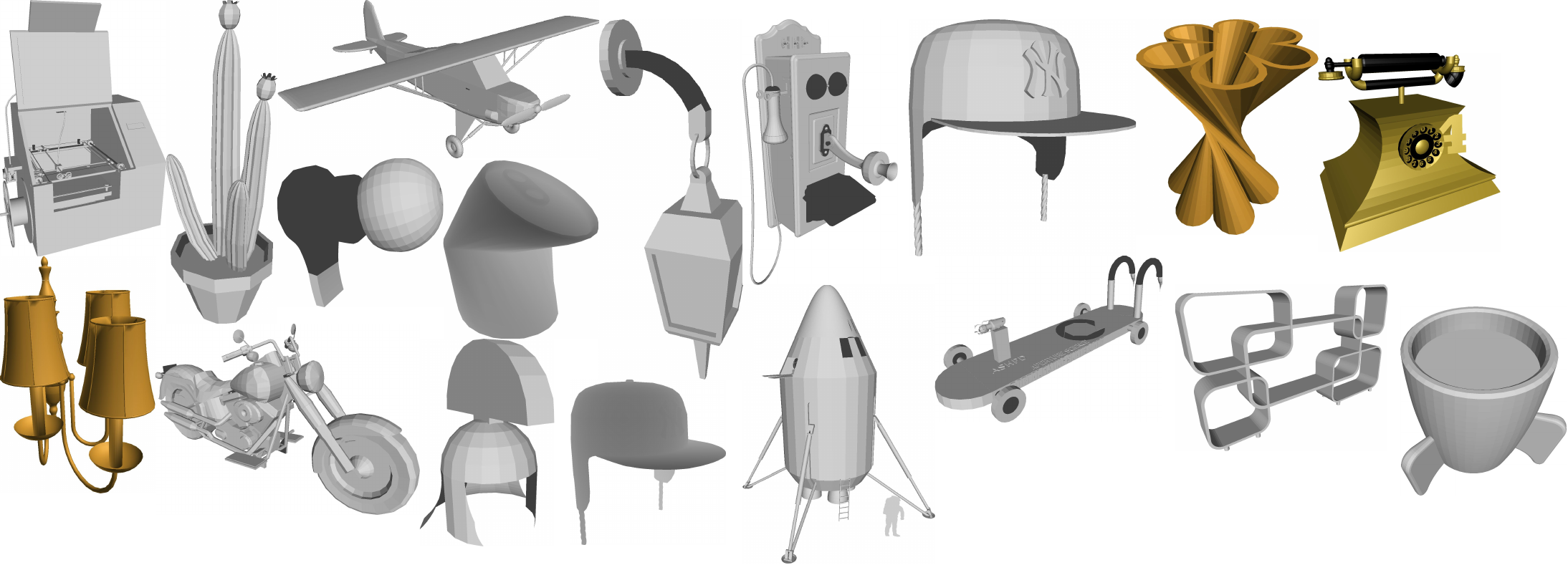}
    }
    \caption{\textbf{Challenge Dataset Meshes} We choose 17 meshes from the ShapeNet~\cite{chang2015shapenet} dataset that feature self-occlusions, asymmetrical geometry, or deviate significantly from the geometry of the YCB~\cite{calli2015ycb} and Grasp Database~\cite{bohg2014data} datasets. }\label{fig:challenge_dataset}

\end{figure} 

All training and evaluation was performed using 590 meshes from the Grasp Database~\cite{bohg2014data} dataset and 28 meshes from the YCB~\cite{calli2015ycb} dataset. 100 meshes were sampled from both datasets to provide 50 validation meshes and 50 test meshes. The validation meshes were used to evaluate the Jaccard quality of each completion whereas the test meshes were used to evaluate the performance of each CNN model. The CNN model that performed best using validation meshes was used for evaluation. We have also created a sample of 17 meshes from the Shapenet~\cite{chang2015shapenet} dataset resized to fit within a grip width of $100mm$. These 17 meshes are not observed during training and are chosen to be difficult to complete with only one view, thus we call this dataset the \textit{Challenge} dataset. All meshes from the challenge dataset are shown in Figure~\ref{fig:challenge_dataset}. All views are voxelized using binvox~\cite{binvox}~\cite{nooruddin03}.

Test object reconstruction results are shown in Table \ref{tab:reconstruction_quality_tests} and challenge reconstruction results are shown in Table \ref{tab:reconstruction_quality_tests_challenge}. The most significant result is the Jaccard and Hausdorff for both the test object and the challenge datasets, with our next-best-view method having the best performance in both. They showed a Jaccard reconstruction of $0.866$ for our next-best-view implementation versus $0.648$ for single-view for unseen objects. The same-view method had demonstrably worse performance than our next-best-view algorithm, with a grasp joint error difference of $32.9\%$, showing that adding weights did not greatly improve performance. Our grasp joint error of $3.28^\circ$ compares favorably with the single view of $4.52^\circ$. For examples of the next-best-view shape completion improvement using the challenge dataset see Figure \ref{fig:completion_improvement}. 

% \begin{figure*}[t]
%     \centering {
%         \includegraphics[width=0.95\linewidth]{imgs/Three View Completion Example.pdf}
%     }
%     \caption{In our experimentation we found that three views did not generally mean better performance over two views. Shown is an example of a single view, two-view with next-best-view, and three-view with two next-best-views and their worse qualitative performance for the single-view and three-view cases. }\label{fig:threeviewexample}

% \end{figure*} 

\begin{figure}[t]
    \centering {
        \includegraphics[width=0.95\linewidth]{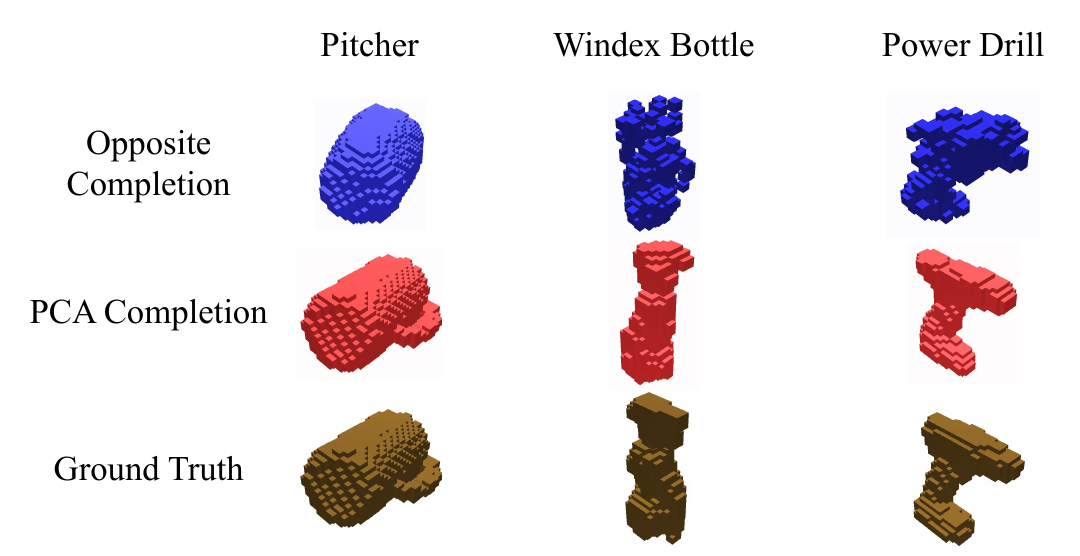}
    }
    \caption{The opposite view is not always enough. Missing an important feature in both the current and the opposite view results in a worse completion and therefore makes it difficult to plan grasps. Utilizing the uncertainty in the single-view completion for next-best-view fixed the missing handle in the reconstruction in the pitcher from the YCB~\cite{calli2015ycb} dataset.}\label{fig:oppositevspca}
\end{figure} 

While three views provided a very small benefit in the Jaccard, Hausdorff, and grasp joint error metrics ($0.868$, $4.782$, and $3.12^\circ$, respectively), the benefit does not warrant the extra effort of capturing a third view in the mobile manipulation step. 
%An example of three views leading to worse performance is shown in Figure~\ref{fig:threeviewexample}. 
Capturing the opposite view, although a reasonable strategy, did not outperform the PCA method. At best, the opposite view performs similarly to the PCA method. At worst, opposite-views miss major features of objects. An example of a completion missing the handle of a pitcher is shown in Figure~\ref{fig:oppositevspca}. Without leveraging the uncertain voxels within the initial completion, naively capturing the opposite view is insufficient for completing the object reliably. In our testing, we found that for the test object dataset, our next-best-view method outperformed the opposite method in all metrics. 

\begin{table}[t]
\vspace{2mm}
    \centering
    \begin{tabular}{|c|c|c|}
        \hline
        \textbf{Localization} & \textbf{Jaccard} & \textbf{Hausdorff}  \\
        \hline
        Registered-Views & 0.923 & 3.523 \\
        \hline
        Next-Best-View & 0.852 & 4.912 \\
        \hline
        Noisy-Views & 0.583 & 9.595 \\
        \hline
    \end{tabular}
    \caption{\textbf{Noisy Odometry Results} We compare the results of our two-view system using PCA versus a noisy odometry model. We find that our system outperforms a model using noisy odometry at runtime to align the two views, justifying the use of a model without registration of the two views for shape completion. }
    \label{tab:noisyodometryresults}
\end{table}

We compared our system versus a version of the single-view model that was trained using two registered views concatenated on the input layer. For training, we overlaid each view using perfect registration. This test is to model how even a small amount of noise in odometry can result in negative performance of a method that relies on idealized registration. We simulated odometry noise by adding up to a $5\%$ error during movement around the pedestal at each time step. We compared both to the Next-Best-View result to evaluate how the noise would compare versus the ideal case and versus a model trained without registration. We found that our model outperformed the noisy case substantially. When overlaying voxel input with translational error the registered-views model was unable to properly complete the mesh. Results are shown in Table \ref{tab:noisyodometryresults}.

\subsection{Navigation Tests}
% \begin{table}

%     \centering
%     \begin{tabular}{|c|c|c|}
%       \hline
%       %This is the header for the table
%       \multicolumn{1}{|c|}{\begin{tabular}[c]{@{}c@{}}\textbf{Next-Best-View} \\  \textbf{Navigation}\end{tabular}}
%       & \multicolumn{1}{c|}{\begin{tabular}[c]{@{}c@{}}\textbf{SPL} \\  \textbf{}\end{tabular}}
%       & \multicolumn{1}{c|}{\begin{tabular}[c]{@{}c@{}}\textbf{Success Rate} \\  \textbf{}\end{tabular}} \\
%       \hline
%       %This is the body for the table
%       ROS (map) & 0.923 & 0.943 \\ \hline
%       Panorama  & 0.854 & 0.845 \\ \hline
%     \end{tabular}

%     \caption{\textbf{Long Range Navigation Results}, measuring the performance of the long range navigation success and path length of two different methods: ROS Navigation Stack and the learning based panoramic target navigation, as described in Section \ref{sec:longrangenavigationtests}. Higher SPL is better.}\label{tab:longrangenavigationtests}
% \end{table}
\begin{table}
    \centering
    \begin{tabular}{|c|c|c|}
      \hline
      %This is the header for the table
      \textbf{Navigation Method} & \textbf{SPL} & \textbf{Success Rate}\\
    %   \multicolumn{1}{|c|}{\begin{tabular}[c]{@{}c@{}}\textbf{Next-Best-View} \\  \textbf{Navigation}\end{tabular}}
    %   & \multicolumn{1}{c|}{\begin{tabular}[c]{@{}c@{}}\textbf{SPL} \\  \textbf{}\end{tabular}}
    %   & \multicolumn{1}{c|}{\begin{tabular}[c]{@{}c@{}}\textbf{Success Rate} \\  \textbf{}\end{tabular}} \\
      \hline
      %This is the body for the table
      ROS Nav (map) & 0.958 & 0.953 \\ \hline
      True Panorama & 0.874 & 0.897 \\ \hline
      Predicted     & 0.813 & 0.853 \\ \hline
    \end{tabular}
    \caption{\textbf{Next-Best-View Navigation Results}, measuring the performance of the next-best-view navigation success and path length compared using ROS, using the true panorama, or a predicted panorama, as described in Section \ref{sec:predicted_view_navigation_tests}. Higher SPL is better.}\label{tab:nextbestviewnavigationresults}
\vspace{-2mm}
\end{table}

% \subsection{Environment Setup}
% \label{sec:environmentsetup}
For testing we used the \textit{house1} environment from the Matterport 3D~\cite{Matterport3D} dataset modified with four tables placed in the house. The Matterport 3D dataset features real world homes scanned and turned into navigable 3D meshes for research. The robot used for navigation and end-to-end testing is the Fetch~\cite{fetch} robot with its parallel jaw gripper. These tables were placed such that the Fetch could adequately navigate around them while still being able to navigate through the environment. We placed different household meshes on these tables to evaluate grasping performance. We reserve 50 test meshes from the YCB~\cite{calli2015ycb} and GRASP~\cite{bohg2014data} datasets not seen during training to be placed randomly on each trial. We used this environment for the navigation and end-to-end testing. Tables were placed in the environment that would accommodate the Fetch's grasp workspace. Table heights were varied between 0.65m, 0.7m, 0.75m, and 0.8m. These heights are high enough off the ground to allow the Fetch to raise and lower its torso to view the object from a variety of different vectors.

\label{sec:predicted_view_navigation_tests}
We validate our agent's ability to navigate locally via two baselines: 1) the \textbf{ROS Navigation Stack} and 2) the learned navigation model using the \textbf{true panorama}. 
The \textbf{ROS Navigation stack}~\cite{ros} uses Dijkstra's algorithm to plan paths and is given a point goal and a global map to get theoretically optimal performance with perfect information. The \textbf{true panorama} is captured at the next-best-view target location to evaluate how the agent performs when given an ideal panorama. Finally we evaluate the performance of our system with the predicted panorama at the next-best-view location. We evaluate the system using Success Weighted by Path Length, or \textbf{SPL}, and \textbf{Success Rate} of reaching the goal location. \textbf{SPL}, from Anderson et al.~\cite{anderson2018evaluation}, is shown in Formula (\ref{spl}), where $l_i$ is the shortest-path distance from the agent's starting position to the goal in episode $i$, $p_i$ is the length of the path actually taken by the agent in this episode, and $S_i$ is a binary indicator of success in trial $i$. $p_i$ is calculated via the L2 distance between each step in the ground truth trajectory using Dijkstra's algorithm. $l_i$ is calculated via the L2 distance between each step in the executed trajectory. This metric weighs each success by the quality of path and thus is always less than or equal to \textit{Success Rate}.
\begin{equation} \label{spl}    
    SPL = \frac{1}{N} \sum_{i=1}^{N} S_i \frac{l_i}{\max (p_i, l_i)}
\end{equation}
We used 50 unseen meshes for our shape completion system. For each object we only consider which orientations, out of the maximum 726, the object rested on the table. Out of 36300 orientations 3455 were stable, and thus we had 3455 sample next-best-views. Table \ref{tab:nextbestviewnavigationresults} shows the results for our method in terms of SPL and success rate. 
We compared to the \textbf{ROS Navigation Stack} knowing it would outperform our method because it has access to a map of the environment. It made for a good baseline with a success rate of navigating to the goal of 0.953. The predicted and ground truth panoramas were both equally effective at navigating to the target goal with a success rate of 0.853 and 0.897, respectively. The effective navigation to predicted panoramic goals validates the use of the predicted panorama in our system. 

\subsection{End-to-End Mobile Manipulation Testing}
\label{sec:end_to_end_testing}

\begin{table}
\vspace{2mm}
    \centering
    \label{tab:longrangenavigationresults}
    \begin{tabular}{|c|c|c|c|c|}
      \hline
      %This is the header for the table
      \multicolumn{1}{|c|}{\begin{tabular}[c]{@{}c@{}}\textbf{Navigation} \\  \textbf{Method}\end{tabular}} 
    & \multicolumn{1}{c|}{\begin{tabular}[c]{@{}c@{}}\textbf{Localized} \\  \textbf{}\end{tabular}} 
    & \multicolumn{1}{c|}{\begin{tabular}[c]{@{}c@{}}\textbf{Completion} \\  \textbf{Method}\end{tabular}} 
    & \multicolumn{1}{c|}{\begin{tabular}[c]{@{}c@{}}\textbf{E2ESPL} \\  \textbf{}\end{tabular}}
    & \multicolumn{1}{c|}{\begin{tabular}[c]{@{}c@{}}\textbf{Grasp} \\  \textbf{Success}\end{tabular}}
      \\
      \hline
      %This is the body for the table
      ROS Nav (map) & Yes & Two-View & 0.892 & 0.884 \\ \hline
      ROS Nav (map) & Yes & Single-View & 0.771 & 0.781 \\ \hline
      Noisy ROS (map) & Yes & Two-View & 0.562 & 0.531 \\ \hline
      True Panorama & No & Two-View & 0.845 & 0.872 \\ \hline
      Predicted     & No & Two-View & 0.820 & 0.819 \\ \hline
    \end{tabular}
    \caption{\textbf{End-to-End Mobile Manipulation Results}, measuring the performance of the full pipeline success rate and E2ESPL with different methods using 50 novel objects. Higher E2ESPL is better. As described in Section \ref{sec:end_to_end_testing}}\label{tab:endtoendresults}
    \vspace{-2mm}
\end{table}

Once we evaluate the major contributions of this work in isolation, we need to evaluate the performance of the end-to-end mobile manipulation system to navigate to a goal and manipulate the target object. Our test trajectories are between 1.5m and 20m in length. For this test, we ablate the completion method and navigation method. We use the \textbf{ROS Navigation Stack} as described in Section \ref{sec:predicted_view_navigation_tests}, the \textbf{True Panorama} method, and our \textbf{predicted panorama} method as navigation modules. We use the \textbf{two-view} and \textbf{single-view} reconstruction methods for these tests. We test single-view and ROS Navigation together then two-view with every navigation method. Additionally, we evaluated the performance with the ROS navigation stack with a noisy odometry model that has imperfect information at runtime about how the robot moves by 5\%. All of our depth information coming from the depth camera have a similar noise model. We performed 400 trials of our system against unseen target locations with 50 novel objects placed on tables in the environment. The SPL metric we used before defined a binary success of a trial $i$ as $S_i$. If we instead replace this with a binary signal of successfully picking up an object $S_p$, we can evaluate the full end to end pipeline. We call this new metric E2ESPL. 
% as whether the agent arrived at the goal location $S_i$. We modify this success signal to be whether the agent successfully picked up the object, or $S_p$. We define a new metric End-to-End Success Weighted by Path Length, or $E2ESPL$, that uses this new success value as the following:
% \begin{equation} \label{e2espl}    
%     E2ESPL = \frac{1}{N} \sum_{i=1}^{N} S_p \frac{l_i}{\max (p_i, l_i)}
% \end{equation}
% \noindent Where $S_p$ is the success of picking up an object, $l_i$ is the shortest path distance from the start location to the end location of trial $i$, $p_i$ is the length of the path taken by the agent, and $N$ is the number of trials. 

Our method was able to reliably navigate to the target positions and grasp the object in $80.7\%$ of our tests despite not having access to a map at runtime or the true panorama of the next-best-view target position. The ROS Navigation stack version was able to navigate to and grasp the object $84.3\%$ of the time and the true panorama was able to navigate $83.6\%$ of the time. Additionally, the single-view completion and ROS Navigation stack succeed $76.9\%$ of the time, showing that two-views helps for unseen objects. We found that our model greatly outperformed ROS with a noisy odometry model where ROS was unable to localize itself with noisy sensors. We found that our method failed mostly when colliding with the environment. Results for the end-to-end testing are in Table \ref{tab:endtoendresults}.
\section{Discussions \& Conclusion}
\label{sec:Conclusion}
We proposed an end-to-end mobile manipulation system that navigates to and manipulates an object without localization using a novel panoramic prediction method. We were able to improve the chance of object manipulation success using a novel two-view reconstruction architecture. We demonstrated that our mobile manipulation system leveraging multiple views performs competitively against a method with perfect odometry and a map. We showed that a next-best-view two-view completion model outperforms single-view reconstructions for unseen objects increasing grasp success with a mobile robot. We observed that the learned navigation system can utilize predicted panoramic targets effectively allowing the agent to generate its own goals. 

We found failure to complete the manipulation task could be attributed to poor reconstruction of unseen objects, failure to navigate to the predicted panoramic goal location, or bumping into the environment during a trial. Additionally, there were times our navigation framework reported done prematurely. In future work, it would improve performance to address the environment collisions by introducing a replanning step if a command would result in collision. Furthermore, it would be useful to handle cases where the agent cannot navigate to the unseen view and find a closest position to navigate to next. Reachability of next-best-views would similarly benefit grasp success. Designing a system to incorporate reachability would also allow for more than one object to be considered for manipulation. Currently the agent is trained and evaluated in the Gibson simulator. When lab access restrictions due to the pandemic are lifted, we would like to test our system on a real physical robot.

\bibliographystyle{IEEEtran}
\bibliography{IEEEabrv,mybib}

\begin{thebibliography}{10}
\providecommand{\url}[1]{#1}
\csname url@rmstyle\endcsname
\providecommand{\newblock}{\relax}
\providecommand{\bibinfo}[2]{#2}
\providecommand\BIBentrySTDinterwordspacing{\spaceskip=0pt\relax}
\providecommand\BIBentryALTinterwordstretchfactor{4}
\providecommand\BIBentryALTinterwordspacing{\spaceskip=\fontdimen2\font plus
\BIBentryALTinterwordstretchfactor\fontdimen3\font minus
  \fontdimen4\font\relax}
\providecommand\BIBforeignlanguage[2]{{%
\expandafter\ifx\csname l@#1\endcsname\relax
\typeout{** WARNING: IEEEtran.bst: No hyphenation pattern has been}%
\typeout{** loaded for the language `#1'. Using the pattern for}%
\typeout{** the default language instead.}%
\else
\language=\csname l@#1\endcsname
\fi
#2}}

\bibitem{kalashnikov2018qt}
D.~Kalashnikov, A.~Irpan, P.~Pastor, J.~Ibarz, A.~Herzog, E.~Jang, D.~Quillen,
  E.~Holly, M.~Kalakrishnan, V.~Vanhoucke, \emph{et~al.}, ``Qt-opt: Scalable
  deep reinforcement learning for vision-based robotic manipulation,''
  \emph{arXiv preprint arXiv:1806.10293}, 2018.

\bibitem{mahler2019learning}
J.~Mahler, M.~Matl, V.~Satish, M.~Danielczuk, B.~DeRose, S.~McKinley, and
  K.~Goldberg, ``Learning ambidextrous robot grasping policies,'' \emph{Science
  Robotics}, vol.~4, no.~26, 2019.

\bibitem{jacobus2015automated}
C.~J. Jacobus, G.~J. Beach, and S.~Rowe, ``Automated warehousing using robotic
  forklifts,'' 2 2015, uS Patent 8,965,561.

\bibitem{schwarz2017nimbro}
M.~Schwarz, T.~Rodehutskors, D.~Droeschel, M.~Beul, M.~Schreiber, N.~Araslanov,
  I.~Ivanov, C.~Lenz, J.~Razlaw, S.~Sch{\"u}ller, \emph{et~al.}, ``Nimbro
  rescue: Solving disaster-response tasks with the mobile manipulation robot
  momaro,'' \emph{Journal of Field Robotics}, vol.~34, pp. 400--425, 2017.

\bibitem{orsag2017dexterous}
M.~Orsag, C.~Korpela, S.~Bogdan, and P.~Oh, ``Dexterous aerial robots—mobile
  manipulation using unmanned aerial systems,'' \emph{IEEE Transactions on
  Robotics}, vol.~33, no.~6, pp. 1453--1466, 2017.

\bibitem{wang2020multi}
C.~Wang, Q.~Zhang, Q.~Tian, S.~Li, X.~Wang, D.~Lane, Y.~Petillot, Z.~Hong, and
  S.~Wang, ``Multi-task reinforcement learning based mobile manipulation
  control for dynamic object tracking and grasping,'' \emph{arXiv preprint
  arXiv:2006.04271}, 2020.

\bibitem{bajcsy2018revisiting}
R.~Bajcsy, Y.~Aloimonos, and J.~K. Tsotsos, ``Revisiting active perception,''
  \emph{Autonomous Robots}, vol.~42, no.~2, pp. 177--196, 2018.

\bibitem{dissanayake2001solution}
M.~G. Dissanayake, P.~Newman, S.~Clark, H.~F. Durrant-Whyte, and M.~Csorba, ``A
  solution to the simultaneous localization and map building (slam) problem,''
  \emph{IEEE Transactions on robotics and automation}, vol.~17, no.~3, pp.
  229--241, 2001.

\bibitem{zhu2017target}
Y.~Zhu, R.~Mottaghi, E.~Kolve, J.~J. Lim, A.~Gupta, L.~Fei-Fei, and A.~Farhadi,
  ``Target-driven visual navigation in indoor scenes using deep reinforcement
  learning,'' in \emph{2017 IEEE international conference on robotics and
  automation (ICRA)}.\hskip 1em plus 0.5em minus 0.4em\relax IEEE, 2017, pp.
  3357--3364.

\bibitem{francis2020long}
A.~Francis, A.~Faust, H.-T. Chiang, J.~Hsu, J.~C. Kew, M.~Fiser, and T.-W.~E.
  Lee, ``Long-range indoor navigation with prm-rl,'' \emph{IEEE Transactions on
  Robotics}, 2020.

\bibitem{mirowski2016learning}
P.~Mirowski, R.~Pascanu, F.~Viola, H.~Soyer, A.~J. Ballard, A.~Banino,
  M.~Denil, R.~Goroshin, L.~Sifre, K.~Kavukcuoglu, \emph{et~al.}, ``Learning to
  navigate in complex environments,'' \emph{arXiv preprint:1611.03673}, 2016.

\bibitem{calli2015ycb}
B.~Calli, A.~Singh, A.~Walsman, S.~Srinivasa, P.~Abbeel, and A.~M. Dollar,
  ``The ycb object and model set: Towards common benchmarks for manipulation
  research,'' in \emph{Advanced Robotics (ICAR), 2015 International Conference
  on}.\hskip 1em plus 0.5em minus 0.4em\relax IEEE, 2015, pp. 510--517.

\bibitem{Matterport3D}
A.~Chang, A.~Dai, T.~Funkhouser, M.~Halber, M.~Niessner, M.~Savva, S.~Song,
  A.~Zeng, and Y.~Zhang, ``Matterport3d: Learning from rgb-d data in indoor
  environments,'' \emph{International Conference on 3D Vision (3DV)}, 2017.

\bibitem{watkins2019learning}
D.~Watkins-Valls, J.~Xu, N.~Waytowich, and P.~Allen, ``Learning your way
  without map or compass: Panoramic target driven visual navigation,''
  \emph{IROS 2020}, 2020.

\bibitem{HERB}
S.~S. Srinivasa, D.~Ferguson, C.~J. Helfrich, D.~Berenson, A.~Collet,
  R.~Diankov, G.~Gallagher, G.~Hollinger, J.~Kuffner, and M.~V. Weghe, ``{HERB:
  A home exploring robotic butler},'' \emph{Autonomous Robots}, vol.~28, no.~1,
  pp. 5--20, 2010.

\bibitem{gofetch}
\BIBentryALTinterwordspacing
K.~Blomqvist, M.~Breyer, A.~Cramariuc, J.~F{\"{o}}rster, M.~Grinvald,
  F.~Tschopp, J.~J. Chung, L.~Ott, J.~Nieto, and R.~Siegwart, ``{Go Fetch:
  Mobile Manipulation in Unstructured Environments},'' pp. 1--4, 2020.
  [Online]. Available: \url{http://arxiv.org/abs/2004.00899}
\BIBentrySTDinterwordspacing

\bibitem{DOMel2017}
A.~{D {\"{O}} Mel}, S.~Kriegel, M.~Ka{\ss}ecker, M.~Brucker, T.~Bodenmuller,
  and M.~Suppa, ``{Toward fully autonomous mobile manipulation for industrial
  environments},'' \emph{International Journal of Advanced Robotic Systems},
  vol.~14, no.~4, pp. 1--19, 2017.

\bibitem{Wu2020}
J.~Wu, X.~Sun, A.~Zeng, S.~Song, J.~Lee, S.~Rusinkiewicz, and T.~Funkhouser,
  ``{Spatial Action Maps for Mobile Manipulation},'' 2020.

\bibitem{anderson2018evaluation}
P.~Anderson, A.~Chang, D.~S. Chaplot, A.~Dosovitskiy, S.~Gupta, V.~Koltun,
  J.~Kosecka, J.~Malik, R.~Mottaghi, M.~Savva, \emph{et~al.}, ``On evaluation
  of embodied navigation agents,'' \emph{arXiv preprint arXiv:1807.06757},
  2018.

\bibitem{bansal2020combining}
S.~Bansal, V.~Tolani, S.~Gupta, J.~Malik, and C.~Tomlin, ``Combining optimal
  control and learning for visual navigation in novel environments,'' in
  \emph{Conference on Robot Learning}.\hskip 1em plus 0.5em minus 0.4em\relax
  PMLR, 2020, pp. 420--429.

\bibitem{lind2018deep}
L.~Lind, ``Deep learning navigation for ugvs on forests paths,'' 2018.

\bibitem{richter2017safe}
C.~Richter and N.~Roy, ``Safe visual navigation via deep learning and novelty
  detection,'' 2017.

\bibitem{relomogen}
\BIBentryALTinterwordspacing
F.~Xia, C.~Li, R.~Mart{\'{\i}}n{-}Mart{\'{\i}}n, O.~Litany, A.~Toshev, and
  S.~Savarese, ``Relmogen: Leveraging motion generation in reinforcement
  learning for mobile manipulation,'' \emph{CoRR}, vol. abs/2008.07792, 2020.
  [Online]. Available: \url{https://arxiv.org/abs/2008.07792}
\BIBentrySTDinterwordspacing

\bibitem{varley2017shape}
J.~Varley, C.~DeChant, A.~Richardson, J.~Ruales, and P.~Allen, ``Shape
  completion enabled robotic grasping,'' in \emph{2017 IEEE/RSJ international
  conference on intelligent robots and systems (IROS)}.\hskip 1em plus 0.5em
  minus 0.4em\relax IEEE, 2017.

\bibitem{dai2017shape}
A.~Dai, C.~Ruizhongtai~Qi, and M.~Nie{\ss}ner, ``Shape completion using
  3d-encoder-predictor cnns and shape synthesis,'' in \emph{Proceedings of the
  IEEE Conference on Computer Vision and Pattern Recognition}, 2017.

\bibitem{litany2018deformable}
O.~Litany, A.~Bronstein, M.~Bronstein, and A.~Makadia, ``Deformable shape
  completion with graph convolutional autoencoders,'' in \emph{Proceedings of
  the IEEE conference on computer vision and pattern recognition}, 2018, pp.
  1886--1895.

\bibitem{watkins2019multi}
D.~Watkins-Valls, J.~Varley, and P.~Allen, ``Multi-modal geometric learning for
  grasping and manipulation,'' in \emph{2019 International Conference on
  Robotics and Automation (ICRA)}, 2019, pp. 7339--7345.

\bibitem{Yang18}
B.~Yang, S.~Rosa, A.~Markham, N.~Trigoni, and H.~Wen, ``Dense 3d object
  reconstruction from a single depth view,'' in \emph{TPAMI}, 2018.

\bibitem{3DNVS}
\BIBentryALTinterwordspacing
K.~Ashutosh, S.~Kumar, and S.~Chaudhuri, ``3d-nvs: {A} 3d supervision approach
  for next view selection,'' \emph{CoRR}, vol. abs/2012.01743, 2020. [Online].
  Available: \url{https://arxiv.org/abs/2012.01743}
\BIBentrySTDinterwordspacing

\bibitem{williams2006gaussian}
O.~Williams and A.~Fitzgibbon, ``Gaussian process implicit surfaces,'' 2006.

\bibitem{krainin2010manipulator}
M.~K, P.~H, X.~R, and D.~F, ``Manipulator and object tracking for in hand model
  acquisition.''

\bibitem{krainin2011autonomous}
M.~Krainin, B.~Curless, and D.~Fox, ``Autonomous generation of complete 3d
  object models using next best view manipulation planning,'' in \emph{2011
  IEEE International Conference on Robotics and Automation}.\hskip 1em plus
  0.5em minus 0.4em\relax IEEE, 2011, pp. 5031--5037.

\bibitem{hermann2016eye}
A.~Hermann, F.~Mauch, S.~Klemm, A.~Roennau, and R.~Dillmann, ``Eye in hand:
  Towards gpu accelerated online grasp planning based on pointclouds from
  in-hand sensor,'' in \emph{2016 IEEE-RAS 16th International Conference on
  Humanoid Robots (Humanoids)}.\hskip 1em plus 0.5em minus 0.4em\relax IEEE,
  2016.

\bibitem{connolly1985determination}
C.~Connolly, ``The determination of next best views,'' in \emph{Proceedings.
  1985 IEEE international conference on robotics and automation}, vol.~2.\hskip
  1em plus 0.5em minus 0.4em\relax IEEE, 1985, pp. 432--435.

\bibitem{pito1995solution}
R.~Pito and R.~K. Bajcsy, ``Solution to the next best view problem for
  automated cad model acquisiton of free-form objects using range cameras,'' in
  \emph{Modeling, simulation, and control technologies for manufacturing}, vol.
  2596.\hskip 1em plus 0.5em minus 0.4em\relax International Society for Optics
  and Photonics, 1995, pp. 78--89.

\bibitem{Callieri2004RoboScanAA}
M.~Callieri, A.~Fasano, G.~Impoco, P.~Cignoni, R.~Scopigno, G.~Parrini, and
  G.~Biagini, ``Roboscan: an automatic system for accurate and unattended 3d
  scanning,'' \emph{Proceedings. 2nd International Symposium on 3D Data
  Processing, Visualization and Transmission, 2004. 3DPVT 2004.}, pp. 805--812,
  2004.

\bibitem{chen2005vision}
S.~Chen and Y.~Li, ``Vision sensor planning for 3-d model acquisition,''
  \emph{IEEE Transactions on Systems, Man, and Cybernetics, Part B
  (Cybernetics)}, vol.~35, no.~5, pp. 894--904, 2005.

\bibitem{gomeznbvplanning}
J.~Vasquez-Gomez and L.~Sucar, ``Next-best-view planning for 3d object
  reconstruction under positioning error,'' vol. 7094, 11 2011, pp. 429--442.

\bibitem{karaszewski2016assessment}
M.~Karaszewski, M.~Adamczyk, and R.~Sitnik, ``Assessment of next-best-view
  algorithms performance with various 3d scanners and manipulator,''
  \emph{ISPRS Journal of Photogrammetry and Remote Sensing}, vol. 119, pp.
  320--333, 2016.

\bibitem{mcgreavy2017next}
C.~McGreavy, L.~Kunze, and N.~Hawes, ``Next best view planning for object
  recognition in mobile robotics.''\hskip 1em plus 0.5em minus 0.4em\relax CEUR
  Workshop Proceedings, 2017.

\bibitem{xie2021unseen}
C.~Xie, Y.~Xiang, A.~Mousavian, and D.~Fox, ``Unseen object instance
  segmentation for robotic environments,'' \emph{IEEE Transactions on
  Robotics}, 2021.

\bibitem{kahn2015active}
G.~Kahn, P.~Sujan, S.~Patil, S.~Bopardikar, J.~Ryde, K.~Goldberg, and
  P.~Abbeel, ``Active exploration using trajectory optimization for robotic
  grasping in the presence of occlusions,'' in \emph{2015 IEEE International
  Conference on Robotics and Automation (ICRA)}.\hskip 1em plus 0.5em minus
  0.4em\relax IEEE, 2015, pp. 4783--4790.

\bibitem{Potthast2011NextBV}
C.~Potthast and G.~S. Sukhatme, ``Next best view estimation with eye in hand
  camera,'' in \emph{IROS 2011}, 2011.

\bibitem{gomez2014}
J.~I. Vasquez-Gomez, L.~E. Sucar, and R.~Murrieta-Cid, ``View planning for 3d
  object reconstruction with a mobile manipulator robot,'' in \emph{2014
  IEEE/RSJ International Conference on Intelligent Robots and Systems}, 2014,
  pp. 4227--4233.

\bibitem{fischler1981random}
M.~A. Fischler and R.~C. Bolles, ``Random sample consensus: a paradigm for
  model fitting with applications to image analysis and automated
  cartography,'' \emph{Communications of the ACM}, 1981.

\bibitem{open3d}
Q.-Y. Zhou, J.~Park, and V.~Koltun, ``{Open3D}: {A} modern library for {3D}
  data processing,'' \emph{arXiv:1801.09847}, 2018.

\bibitem{bernardini1999ball}
F.~Bernardini, J.~Mittleman, H.~Rushmeier, C.~Silva, and G.~Taubin, ``The
  ball-pivoting algorithm for surface reconstruction,'' \emph{IEEE transactions
  on visualization and computer graphics}, vol.~5, no.~4, pp. 349--359, 1999.

\bibitem{igibson}
B.~Shen, F.~Xia, C.~Li, R.~Mart{\i}n-Mart{\i}n, L.~Fan, G.~Wang, S.~Buch,
  C.~D’Arpino, S.~Srivastava, L.~P. Tchapmi, K.~Vainio, L.~Fei-Fei, and
  S.~Savarese, ``igibson, a simulation environment for interactive tasks in
  large realistic scenes,'' \emph{arXiv preprint}, 2020.

\bibitem{lorensen1987marching}
W.~E. Lorensen and H.~E. Cline, ``Marching cubes: A high resolution 3d surface
  construction algorithm,'' in \emph{ACM siggraph computer graphics}, vol.~21,
  no.~4.\hskip 1em plus 0.5em minus 0.4em\relax ACM, 1987, pp. 163--169.

\bibitem{miller2004graspit}
A.~T. Miller and P.~K. Allen, ``Graspit! a versatile simulator for robotic
  grasping,'' \emph{IEEE R\&A Magazine}, vol.~11, no.~4, pp. 110--122, 2004.

\bibitem{ferrari1992planning}
C.~Ferrari and J.~Canny, ``Planning optimal grasps,'' in \emph{Robotics and
  Automation, 1992. Proceedings., 1992 IEEE International Conference on}.\hskip
  1em plus 0.5em minus 0.4em\relax IEEE, 1992, pp. 2290--2295.

\bibitem{sucan2013moveit}
I.~A. Sucan and S.~Chitta, ``Moveit!'' \emph{http://moveit.ros.org}, 2013.

\bibitem{jaccard}
S.~Kosub, ``A note on the triangle inequality for the jaccard distance,''
  \emph{arXiv:1612.02696}, 2016.

\bibitem{huttenlocher1993comparing}
D.~P. H, G.~A. K, and W.~J. R, ``Comparing images using the hausdorff
  distance,'' \emph{IEEE Transactions on pattern analysis and machine
  intelligence}, vol.~15, no.~9, pp. 850--863, 1993.

\bibitem{chang2015shapenet}
A.~X. Chang, T.~Funkhouser, L.~Guibas, P.~Hanrahan, Q.~Huang, Z.~Li,
  S.~Savarese, M.~Savva, S.~Song, H.~Su, \emph{et~al.}, ``Shapenet: An
  information-rich 3d model repository,'' \emph{arXiv preprint
  arXiv:1512.03012}, 2015.

\bibitem{bohg2014data}
J.~Bohg, A.~Morales, T.~Asfour, and D.~Kragic, ``Data-driven grasp
  synthesis—a survey,'' \emph{Robotics, IEEE Transactions on}, vol.~30,
  no.~2, pp. 289--309, 2014.

\bibitem{binvox}
P.~Min, ``binvox,'' {\tt http://www.patrickmin.com/binvox} or {\tt
  https://www.google.com/search?q=binvox}, 2004 - 2019, accessed: 2021-09-14.

\bibitem{nooruddin03}
F.~S. Nooruddin and G.~Turk, ``Simplification and repair of polygonal models
  using volumetric techniques,'' \emph{IEEE Transactions on Visualization and
  Computer Graphics}, vol.~9, no.~2, pp. 191--205, 2003.

\bibitem{fetch}
M.~Wise, M.~Ferguson, D.~King, E.~Diehr, and D.~Dymesich, ``Fetch and freight :
  Standard platforms for service robot applications,'' 2016.

\bibitem{ros}
\BIBentryALTinterwordspacing
{Stanford Artificial Intelligence Laboratory et al.}, ``Robotic operating
  system.'' [Online]. Available: \url{https://www.ros.org}
\BIBentrySTDinterwordspacing

\end{thebibliography}

\end{document}